# Named Entity Recognition System for Sindhi Language: First International Conference, iCETiC 2018, London, UK, August 23–24, 2018, Proceedings

Chapter · July 2018

DOI: 10.1007/978-3-319-95450-9_20

# Named Entity Recognition System for Sindhi Language

Awais Khan Jumani[1(✉)], Mashooque Ahmed Memon[2],
Fida Hussain Khoso[3], Anwar Ali Sanjrani[4], and Safeeullah Soomro[5]

[1] Department of Computer Science, Shah Abdul Latif University,
Khairpur Mirs, Pakistan
`awaisjumani@yahoo.com`
[2] Department of Computer Science, Benazir Bhutto Shaheed University,
Layari, Karachi, Pakistan
`pashamorai786@gmail.com`
[3] Department of Basic Sciences, Dawood University of Engineering
and Technology, Karachi, Pakistan
`fidahussain.khoso@duet.edu.pk`
[4] Department of Computer Science, University of Baluchistan, Quetta, Pakistan
`anwar.csd@gmail.com`
[5] Department of Computer Science, AMA International University,
Salmabad, Bahrain
`s.soomro@amaiu.edu.pk`

**Abstract.** Named Entity Recognition (NER) System aims to extract the existing information into the following categories such as: Person's Name, Organization, Location, Date and Time, Term, Designation and Short forms. Now, it is considered to be important aspect for many natural languages processing (NLP) tasks such as: information retrieval system, machine translation system, information extraction system and question answering. Even at a surface level, the understanding of the named entities involved in a document gives richer analytical framework and cross referencing. It has been used for different Arabic Script-Based languages like, Arabic, Persian and Urdu but, Sindhi could not come into being yet. This paper explains the problem of NER in the framework of Sindhi Language and provides relevant solution. The system is developed to tag ten different Named Entities. We have used Ruled based approach for NER system of Sindhi Language. For the training and testing, 936 words were used and calculated performance accuracy of 98.71%.

## 1 Introduction

During our past school days it is being taught us that a proper noun is a "specific person, place, or thing," so this definition has been taken from a concrete noun. Unfortunately, it is observed that simple mnemonic and computational linguistic tasks are extremely complex, the retrieval system of named entities like that Person's Name, Organization, Location, Date and Time, Term, Designation and Short Term. Actually, the classification of named entity system can be termed as the identification of Named Entities in computer understandable text through information retrieval system can be categories with annotation [1, 2].

It is not only observed that information retrieval system is a subtask of NER but it can play a vital role for reference resolution, categories of disambiguation, and meaning representation with many of other NLP applications. Parts of speech tagging, semantic parsers, and thematically meaning representation can be enlarged with tagging system to achieve a better results. On the other hand, specific application of NER system exist in large amount of question answers system, automatic forwarding content, textual requirement and news searching. Even the understanding of NER system provides a better platform for analytical frameworks and cross-referencing. Named entity contains top three level categorizations according to Defense Advanced Research Projects Agency's message to understand the approach of named entities, temporary expressions and number of expressions but the categories of named entities can be described as a unique identifiers of persons' names, locations, events and organizations, it can be considered as entities and a lot of others.

### 1.1 Gazetteer Based Approach

NER System provides many annotations to candidates, and assure to a certain amount, can list the probability of a candidate which is joining to a group or sub-group of a NER. But this is not required as a complete solution of machine learning methods, some knowledge is required for untrained candidacy tokens. Furthermore classification of candidates is required as other issues or problems can be resolved with gazetteer based approach. Gazetteer based approach should be developed to supply external knowledge for learners, or changed to supply unannotated data with their training material.

Therefore, the researchers have come towards the development of gazetteers based encyclopedia of named entities, otherwise, some special applications can be developed on the basis of gazetteer. The Systems can be defined in [3–5] usages of a combinational rule-based devices, parts of speech (POS) tags, and some word frequency can be analyzed to propose these candidates who have no any approach to learn machine learning methods.

## 2 Literature Review

Many researchers have been working with information extraction during my research for the NER system, I have selected this kind of literature. Some of the researchers just give the results of their respective research regarding NER.

Shrimad Hinal, [6] has focused on the NLP, which is being used in different Indian languages and also compared that language with each other through conditional random fields. So, they have proven that which is better for Indian languages to extract the named entity information. Moreover Tarek, [7] has introduced the new method of extraction information for Arabic languages from the news articles. They have been using the two methods for extracting information (RenA and ALDA) which is better than previous tools, for such kind of these methodologies they have been taking accurate results to extract Name, Organization, and Location from online resources. Also Nita, [8] has surveyed that NER system used in different Indian languages and non-Indian languages. They observed that different kinds of NER especially in Indian languages, which techniques and approaches are best for Indian languages. Similarly Ridong, [9] has recognized that many researchers have developed different kind of NER system, it is quite difficult to which is best NER system for new user. So, they have constructed the hybrid NER system for our interest. Likewise, Maithilee, [10] has researched that different type of named entity has been introduced with different languages, this study shows (NERC) NER and classification. Many of the researchers just used rule based approaches which is to perform and this study related to learning based approaches. As well as, Seth, [11] has explored the ways and limitations of data extraction in NER and termed recognition for getting meaningful concept. Also they perform digital humanities research in searching and browsing operations. It is understanding the value of NER system. Correspondingly, Khaled, [12] has described the recent activities and growth regarding Arabic NER study and the importance of Arabic NER characteristics of languages are highlighted. Mainly common tools features can be used in Arabic NER and illustrated the evaluation of their classification. Respectively, Maksim, [13] has explored many combinations of NER features and compared the performance with each other. So, they have built conditional based approach and collected the results, statistical importance of their boost performance with their previous top performance system. Similarly, Sherief, [14] has described the evolution of (NERA) NER system in Arabic language, also focusing the integrating machine learning with rule based approach for NERA. They have implemented the methods with another taking the results regarding NERA and collect best approach with rule based approach and machine learning system. Likewise, Ronan, [15] has discovered the flexibility of NLP in their specific task of engineering and considering a lot of their prior information. Each task can be measured and optimized the features of NLP, so, this system can acquire the internal representation on the basis of huge amount of unlabeled data sets. Also, Darvinder, [16] has surveyed that NER system used in various languages like Chinese and Spanish and so on. In English language a lot of work has been done specially capitalization is more important part of NER system. Secondly, in India Punjabi is official language of Punjab and many of the tagging and information extraction work has been done in it. Respectively, Wenhui, [17] has implemented the semi- supervised algorithm learning methodology with conditional random fields for NER, also this algorithm used better efficiency and redundant the data. It has improved algorithm for the next iteration. Correspondingly, Alireza, [18] has recognized the named entity for extracting the information like person, organization and so on. They have compared the portable message understanding conference which is highly used everywhere, it can be used robust and novel learning based with fuzzy technology. Furthermore, Trian, [19] has presented the experimental results which has been taken with help of support vector machine and applied on Vietnamese language. Through the comparison of conditional random fields, the support vector machine gives better results as compared to CRF. The identification and classification of proper nouns in plain text is of key importance in numerous NLP applications. It is the first step of a desktop application as proper names generally carry important information about the text itself, and thus are targets for extraction. Moreover Sindhi NER (SNER) can be a stand-alone application. It includes proper nouns, dates, identification numbers, phone numbers, e-mail addresses and so on.

## 3   Problem in NER System

Sindhi language is part of Asian Language and many other Asian Languages being a part of this language and they do not need any concept of capitalization. It is noticed European Language that in English this kind of feature is commonly used to identify Named Entity in text therefore all the names of European languages are always start with capital letter. Deficiency of capitalization tool makes the NER task more for Sindhi Language. Sindhi Names consist of lot of confusion so it can be used as a proper noun or common noun. The actual goal of any NER system is to separate or remove proper noun in place of common noun. For instance: شفقت (Shafqat), رحمت (Rehmat), حکمت (Hikmat) or سداقت (Sadaqat) can be counted as a person name or it can be counted as a place. Many other problems are being occurred in Sindhi Language as a standardization of Sindhi spelling as well as word. Multiple spelling formats are available for one Sindhi word. For example tablet can be written in Sindhi with different ways like (ٽڪي ،گوري) (Gori/ Tiki). In another example word like House can be written in with different ways like (جڳھ ،گھر) (Ghar/ Jagah) etc. It is too difficult task for NER System. The Approach of any language is more important for any resources either it is Statistical or Rule based. As consider these kind of problems in Sindhi Language there is no any mechanism or any Gazetteer and annotated data available in Sindhi Language.

Limitation of Sindhi NER system, we analyze some of the fundamental design challenges and misconceptions that underlie the development of an efficient and robust Sindhi NER system. Rule based systems are usually best performing system but suffers some limitation such as language dependent, difficult to adapt changes.

## 4   Rule Based Approach

Sindhi text can be identified by means of these different Rules.

1. Some of the rules should be applied for the better recognition for date and time tags. Such kinds of tag can easily be recognized by regular expressions as it may be generated for particular forms like 05.06.2016 or 05/06/2016 and it is better known as 10:40 or 02:30. The entire system has capability to find the date like 07 جوالء 2016 or 07 جوالء and سال 2016.
2. Many of the locations' names and terms may be used with different identifications for suffix matching. In Sindhi Language and some of the other Asian languages that contain many locations which end on "PUR" just like (Khairpur, Ranipur), "STAN" (Baluchistan, Afghanistan) "GARR" (Muzafargarr, Khangarr), "NAGAR" (Shantinagar, Naseemnagar) and some cities that end on "ABAD" (Islamabad, Nooriabad). Persons' names, Terms and Organizations can be used by Suffix Matching, Just Like:

Persons' name that end on "DAD" (Allahdad, Saindad), "ALLAH" (Hidayatullah, Naimatullah). Some Terms that end on "YAT" (Hayatyat, Falkyat), Some Person's last name ends with "Hassan" or "Hussain" after that, we can identify them as Organization or it may be Persons' names like Zahid Hussain and Ali Hassan.

3. This system uses most common Gazetteer based approach for the common Persons' names for their identification. This system has capability to tag the words of three length as counted one Named Entity just like: محمد علي جماڻي (Muhammad Ali Jumani). Here Form implementation of the Named Entity System we have composed 10000 Sindhi names and 7000 Urdu persons' names.

4. Some of Sindhi and Punjabi Surnames have been stored in this system for the better recognition of the Named Entity, the system can search out His/Her first name, Just like surname (Jumani) جماڻي the system has ability find out the word before the Surname it may be the first name of person like (Awais Jumani). اويس جماڻي.

5. Title of the person can help to find out the designation of people just like وزير اعظم (Wazir-e-Azam) and مسز (Mrs) that contains proper name next to it as to help of searching the title and surname of the person. The system has ability to detect Persons' Named Entity easily and also recognize those names which are not included as a part of gazetteer. We have collected 60 Title Persons and 200 Designations.

6. During the implementation of NLP, our system can easily resolve many problems which occurred for identity of true person. Many of the rules are applied, for instance: the system can find out the opacity in persons' names. So system finds out the word اويس (Awais), then it can easily solve the problem of sentence structure or detect either it is noun or preposition where it may search Title, Person or Designation. If there is no clue to recognize as Named Entity then at last it checks out the post or position of ambiguity word like اويس جي گهر ۾ ڪتاب پيو آهي (Awais je ghar me kitab payo ahe.) the word (je) جي indicates that it is persons' name, Hence the system tag it as Person Named Entity (PNE).

7. One of the rules is to detect the numbers that are not numerals just like سيون، ايٽ، نائن (Seven, Eight, and Nine). Our system can calculate three words as One Number Entity just like شهسوپنج (Six Hundred Five).

8. Persons' Name can be written/searched by the abbreviations just like مهر جي اي (J A Mahar). The system have should search out surname only then it should continuously try to find out the short form of persons' name.

9. This system also have the ability to find out the short form tags like (DDR) ڊي ار ڊي (KTN) ڪي ٽي اين etc.

10. Organizations can be tagged where gazetteer based approach is used. During the testing of our system, we have stored a lot of related organizations' data and some heuristics to search ORG: tags. Suppose, if text includes Org: (Uho Sindh university me parhe tho.). اوهو سند يونيورسٽي ۾ پڙهي ٿو and it does not found out this Organization in gazetteer. The system applies rules to search and tag org: as "Sindh University".

## 5   Flowchart of NER System

The Approach of any language is more important for any resources either it is Statistical or Rule based. Our total works depend upon the given below diagram Fig. 1 shows the flow of working Sindhi NER System.

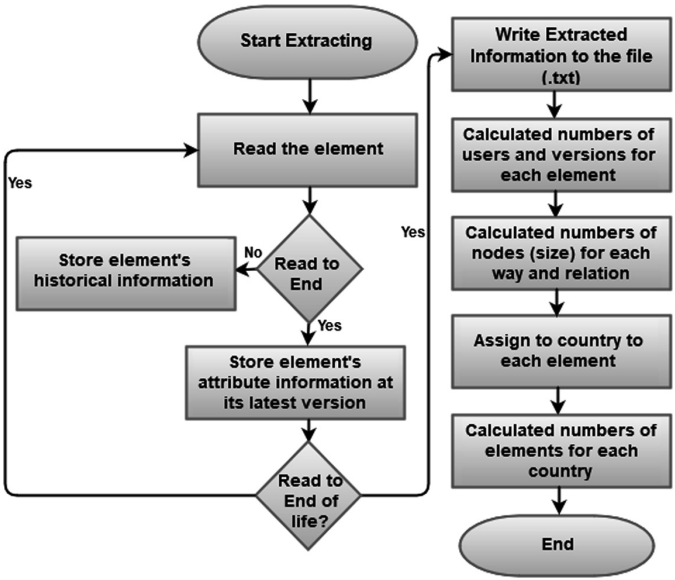

**Fig. 1.** Shows framework of Sindhi NER system

## 6   Algorithm of NER System

The entire system has been developed on visual studio.net platform through the implementation of latest features of named entity, it uses different classes and C sharp programming just like: linked list technique and tokenize the classes and functions.

1. Input text, or file uploading
2. Standard Control of Input Text
    1.1 Eradicate unwanted space
    1.2 Eradicating the special characters from the end of strings.
3. Gazetteer based search
    3.1 Gazetteer based search can be Places, Brands, Abbreviations, Terms and Organizations Tags.
4. Tokenized and Standardized
    4.1 Each word can be tokenized and searched regarding Gazetteer.
5. Searching tags of Date and Time
    5.1 Numeral numbers can be searched and also Date, Time, URL and Email tags.
6. Rules of Person Name tags

- 6.1 It can detect the persons' names with Title, Surname and Designation without any Usage of Gazetteer Based Technique.
7. Removing the Suffix rules
    - 7.1 Removable of suffixes rules it can detect the places' names, organizations, Izaafats and another types of persons' names
8. Searching the names of persons and numbers
    - 8.1 Gazetteer Based can search many of the persons' names with various techniques
    - 8.2 It can be applied for Persons' names equal to three words writing.
    - 8.3 Short form of names can be detected.
    - 8.4 Ambiguity in names can be resolved.
    - 8.5 Numbers can be found in non-numeral form.
9. Some Abbreviations which were not found in the Gazetteer Based Lookup.
10. Organizations Searching Rules
    - 10.1 Some organizations which are not be possible to search in Gazetteer Based Technique, the entire rule can be found and tagged them.
11. Output of the tagged and untagged data

This algorithm is self-explanatory, where some of the steps which are concerned with Gazetteer lookup to extract the various Named Entities in each text. Collection of several Named Entities related to different fields like legislation, commerce and etc. In algorithm's step 3 gazetteer look up can be concerned with Locations, Brands, Terms, Abbreviations and Organizations Tags. These tags are not ambiguous that's why the system can easily work without any rule. Many persons' names have patterns just like suffixes or prefixes of the word, so, the step 6 has capability to detect the same kind of words. It has been taken ideas from another researchers [19].

## 7 Data Collection

In the computer age or digital world, it is almost common to collect data through two sources—primary sources and secondary sources. The data collect through the primary sources is called as primary data; and the data collected through the secondary sources is termed as secondary data according to experts of this field.

Primary Data is also said to be as 'raw data'. This data is actually collected by means of genuine source in a controlled or uncontrolled environment. It means that a controlled environment is based upon experimental research in which the researcher directly controls some variables. On the other hand, data collected by means of observations or questionnaire survey in a natural–cum–practical settings is good example of the data obtained in an uncontrolled environment [6].

Whereas, Secondary Data availed through the secondary sources such as: reports, journals magazines, books, documents, research papers, articles, dictionaries – soft and hard copies and websites etc. The simplest method to guess either a typical phrase is a named entity or not simply to look it up in a gazetteer. Look-up symptoms work prettily only with large entity lists. In case of ambiguous entities, the approach is usually competitive against machine learning algorithms. Generally, in machine learning approaches, Gazetteer features are also most common and performance of identification systems can be further more developed and progressed gradually. Presently, in the computer world and much striving digital world where all living and non-living things are shaping themselves in accordance with the customs of global village, there are a lot of websites resources that are, with less efforts, adaptable and accessible to NER, for instance: Britannica, Wikipedia, video libraries, programing, software and encyclopedia. Surely which are helpful and useful to materialize any dream dreamt by any outstanding and knowledge full person. Even, there will be so many other supplementary digital sources to assist global man in this digital world in coming years [7].

## 8 Layout of Sindhi NER System

Sindhi NER System shows the following extraction and it is connected to backend database. In this system user can search and mark the given below tags so our system will show their respective data Fig. 2 shows the dashboard of Sindhi NER System.

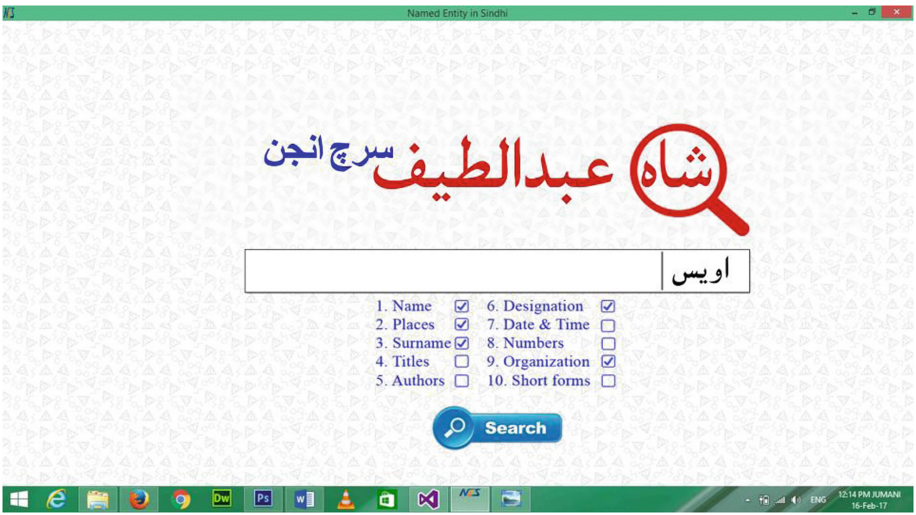

**Fig. 2.** Illustrates the Sindhi NER dashboard

Figure 3 shows the searching results using their tags like name, places, surname, designation and organization. This is the first Sindhi NER system which work like a google search engine and this is the desktop application. User can easily find out their desired data from Sindhi NER system.

**Fig. 3.** Shows the searching results

## 9   Conclusion

NER System aims to extract the existing information into the following categories such as: Person's Name, Organization, Location, Date and Time, Term, Designation and Short forms. Now, it is considered to be important aspect for many natural languages processing tasks such as: information retrieval system, machine translation system, information extraction system and question answering. Even at a surface level, the understanding of the named entities involved in a document gives richer analytical framework and cross referencing. It has been used for different Arabic Script-Based languages like, Arabic, Persian and Urdu but, Sindhi could not come into being yet. This paper explains the problem of NER in the framework of Sindhi Language and provides relevant solution. The system is developed to tag twelve different Named Entities. We have used Ruled based approach for NER system of Sindhi Language. For the training and testing, 936 words were used and calculated performance accuracy of 98.71%. It is a desktop application which recognize the words from database and in future we will work on web application using support vector machine (SVM) approach.

## References

1. Shah, H., Bhandari, P., Mistry, K., Thakor, S., Patel, M., Ahir, K.: Study of NER for Indian languages. Int. J. Inf. Sci. Tech. (IJIST) **6**(1/2), 15–20 (2016)
2. Kanan, T., Ayoub, S., Saif, E., Kanaan, G., Chandrasekar, P., Fox, E.A.: Extracting named entities using named entity recognizer and generating topics using Latent Dirichlet allocation algorithm for Arabic news articles. In: Proceedings of the International Computer Sciences and Informatics Conference (ICSIC) (2016)

3. Patil, N., Patil, A.S., Pawar, B.V.: Survey of NER systems with respect to Indian and Foreign languages. Int. J. Comput. Appl. **134**(16), 21–26 (2016)
4. Jiang, R., Banchs, R.E., Li, H.: Evaluating and combining NER systems. In: Proceedings of the Sixth Named Entity Workshop, Joint with 54th ACL, Berlin, Germany, pp. 21–27 (2016)
5. Patawar, M.M.L., Potey, M.M.A.: Approaches to NER: a survey. Int. J. Innov. Res. Comput. Commun. Eng. **3**(12), 37–42 (2015)
6. van Hooland, S., Wilde, M.D., Verborgh, R., Steiner, T., Van de Walle, R.: Exploring entity recognition and disambiguation for cultural heritage collections. J. Digital Sch. Humanit. **30**, 262–279 (2014)
7. Shaalan, K.: A survey of Arabic NER and classification. Assoc. Comput. Linguist. **40**(2), 469–510 (2014)
8. Tkachenko, M., Simanovsky, A.: NER: exploring features. In: Proceedings of KONVENS 2012, Vienna (2012)
9. Abdallah, S., Shaalan, K., Shoaib, M.: Integrating Rule-Based System with Classification for Arabic NER, pp. 311–322. Springer-Verlag, Heidelberg (2012)
10. Collobert, R., Weston, J., Bottou, L., Karlen, M., Kavukcuoglu, K., Kuksa, P.: Natural language processing (Almost) from scratch. J. Mach. Learn. Res. **12**, 2493–2537 (2011)
11. Kaur, D., Gupta, V.: A survey of NER in English and other Indian languages. Int. J. Comput. Sci. Issues **7**(6), 89–95 (2010)
12. Liao, W., Veeramachaneni, S.: A simple semi-supervised algorithm for NER. In: Proceedings of the NAACL HLT Workshop on Semi-supervised Learning for Natural Language Processing, pp. 58–65 (2009)
13. Mansouri, A., Affendey, L.S., Mamat, A.: NER approaches. Int. J. Comput. Sci. Netw. Secur. **8**(2), 67–71 (2008)
14. Tran, T., Pham, T., Hung, T.X., Dinh, D., Collier, N.: NER in Vietnamese documents. Natural Institute of Informatics (2007)
15. Singh, U.P., Goyal, V., Lehal, G.S.: NER system for Urdu. In: Proceedings of Cooling Mumbai, pp. 2507–2518 (2012)
16. Kazama, J., Torisawa, K.: Exploiting Wikipedia as external knowledge for NER. In: Joint Conference on Empirical Methods in Natural Language Processing and Computational Natural Language Learning, pp. 698–707 (2007)
17. Alexander, E., Richman, P., Schone, P.: Mining Wiki resources for multilingual NER. In: Proceedings of the 46th Annual Meeting of the Association of Computational Linguistics: Human Language Technologies, Stroudsburg, PA, pp. 1–9 (2008)
18. Nadeau, D., Sekine, S.: A survey of NER and classification (2008). http://nlp.cs.nyu.edu/sekine/papers/li07.pdf
19. Belgaum, M.R., Soomro, S., Alansari, Z., Alam, M.: Ideal node enquiry search algorithm (INESH) in MANETS. Ann. Emerg. Technol. Comput. (AETiC) **1**(1), 26–33 (2017)